\documentclass[conference]{IEEEtran}
\IEEEoverridecommandlockouts
\usepackage{cite}
\usepackage{amsmath,amssymb,amsfonts}
\usepackage{algorithmic}
\usepackage{graphicx}
\usepackage{textcomp}
\usepackage{xcolor}
\usepackage{hyperref}
\def\BibTeX{{\rm B\kern-.05em{\sc i\kern-.025em b}\kern-.08em
    T\kern-.1667em\lower.7ex\hbox{E}\kern-.125emX}}
\begin{document}

\title{OpenICS: Open Image Compressive Sensing Toolbox and Benchmark
}
\author{\IEEEauthorblockN{Jonathan Zhao\textsuperscript{*}, Matthew Westerham\textsuperscript{*}, Mark  Lakatos-Toth\textsuperscript{*}, Zhikang Zhang\textsuperscript{*}, Avi Moskoff, Fengbo Ren\thanks{* equally contribute}}
\IEEEauthorblockA{\textit{Parallel Systems and Computing Laboratory}  \\
\textit{Arizona State University}\\
Tempe, USA \\
jjzhao@asu.edu, mkweste1@asu.edu, mlakato1@asu.edu, zzhan362@asu.edu, amoskoff@asu.edu, renfengbo@asu.edu}\\
}

\maketitle

\begin{abstract}
We present OpenICS, an image compressive sensing toolbox that includes multiple image compressive sensing and reconstruction algorithms proposed in the past decade. Due to the lack of standardization in the implementation and evaluation of the proposed algorithms, the application of image compressive sensing in the real-world is limited. We believe this toolbox is the first framework that provides a unified and standardized implementation of multiple image compressive sensing algorithms. In addition, we also conduct a benchmarking study on the methods included in this framework from two aspects: reconstruction accuracy and reconstruction efficiency. We wish this toolbox and benchmark can serve the growing research community of compressive sensing and the industry applying image compressive sensing to new problems as well as developing new methods more efficiently. Code and models are available at \url{https://github.com/PSCLab-ASU/OpenICS}. The project is still under maintenance, and we will keep this document updated.
\end{abstract}

\section{Introduction}
Compressive sensing is a signal sensing technique that performs the sensing and compression of signals simultaneously to reduce the sensing cost without losing information. Over the past decade, there is a wide variety of image compressive sensing reconstruction methods proposed. However, due to the lack of standardization in the implementation and evaluation of the proposed algorithms, the application of image compressive sensing in the real-world is still limited. Towards the goal of efficient deployment and evaluation of image compressive sensing, we build OpenICS which is an image compressive sensing toolbox containing multiple image compressive sensing reconstruction methods implemented in a unified interface and structure. 

Major features of OpenICS are 1. \textbf{Unified interface.} We rewrite the code of multiple image compressive sensing algorithms to build a unified interface for all the methods. This unified design greatly improves the usability and availability of our toolbox. 2. \textbf{Modular design.} Each method locates in a separate folder, and there is no cross-dependency between different methods. This modular design improves the reusability of our toolbox. 3. \textbf{Out-of-the-box usage.} Our toolbox contains the most representative methods of image compressive sensing methods. See Section 2 for the full list. 

In addition to the toolbox we implemented, we also propose a benchmark to evaluate all the methods included in the toolbox from two aspects: reconstruction accuracy and reconstruction speed. The benchmark results include the performance of all the methods evaluated on six different datasets and five different compression ratios. We believe our benchmark is the most complete benchmark in the domain of image compressive sensing so far in terms of the variety of datasets and the range of compression ratios. 

Our contribution are summarized as follows:

1. We provide a toolbox in the domain of image compressive sensing that consists of multiple most representative algorithms in this domain. The toolbox has a unified interface, modular design, and it can be used out of box. 

2. We propose a benchmark in the domain of image compressive sensing and use it to evaluate the methods included in our toolbox. It is by far the most complete benchmark in the domain of image compressive sensing in terms of the variety of datasets and the range of compression ratios.

\section{Methods Included}
OpenICS contains implementations of multiple image CS reconstruction methods. Based on whether the method is data-dependent, we divide implemented methods into two categories: Model-based methods and data-driven methods.  
\subsection{Model-based Methods}
Model-based methods use pre-defined models based on prior knowledge of the signals to perform the reconstruction. The included model-based methods are listed and summarized in table~\ref{methodlist}.

\textbf{L1\cite{l1}}: The first reconstruction methods in the domain of compressive sensing(Only the total-variation-based methods are currently implemented).

\textbf{NLR-CS\cite{nlrcs}}: A reconstruction method based on non-local
low-rank regularization.

\textbf{TVAL-3\cite{tval3}}: An efficient image reconstruction method based on total variation minimization.

\textbf{D-AMP\cite{damp}}: An reconstruction method based on model-based image denoising algorithms. 
\subsection{Data-driven Methods}
Data-driven methods do not rely on pre-defined models of signals. Instead, they use neural networks to model the images and perform the reconstruction tasks. The included data-driven methods are listed below.

\textbf{ReconNet\cite{reconnet}}: An end-to-end reconstruction network based on convolutional neural networks. 

\textbf{LDAMP\cite{ldamp}}: An end-to-end reconstruction network built from the unrolled iterative image denoising process by replacing the model-based image denoisers with neural-network-based denoisers.

\textbf{ISTA-Net\cite{istanet}}: An end-to-end reconstruction network built by unrolling the conventional iterative shrinkage-thresholding algorithm. 

\textbf{LAPRAN\cite{lapran}}: An end-to-end reconstruction network based on deep laplacian pyramid neural networks.

\textbf{CSGM\cite{csgm}}: An iterative reconstruction method based on generative adversial neural network. 

\textbf{CSGAN\cite{csgan}}: A variant of CSGM method enhanced by meta-learning to improve reconstruction speed.  

% A table to illustrate the difference of methods from three aspects: 1. requirement of data(yes/no) 2. way to reconstruct(end-to-end/iterative) 3. running platform(cpu/gpu)
\begin{table}[]
\centering
\begin{tabular}{|l|l|l|l|}
\hline
Methods  & Data dependent & Running process & Platform \\ \hline
L1       & No              & Iterative              & CPU              \\ \hline
TVAL-3   & No              & Iterative              & CPU              \\ \hline
NLR-CS   & No              & Iterative              & CPU              \\ \hline
D-AMP    & No              & Iterative              & CPU              \\ \hline
ReconNet & Yes             & End-to-end             & GPU              \\ \hline
ISTA-Net & Yes             & End-to-end             & GPU              \\ \hline
LDAMP    & Yes             & End-to-end             & GPU              \\ \hline
CSGM     & Yes             & Iterative              & GPU              \\ \hline
LAPRAN   & Yes             & End-to-end             & GPU              \\ \hline
CSGAN    & Yes             & Iterative              & GPU              \\ \hline
\end{tabular}
\vspace{4mm}
\caption{List of methods included in OpenICS}
\label{methodlist}
\end{table}

\section{Architecture}
\subsection{Toolbox Structure}
There are two programing languages used to implement all the methods. L1, NLR-CS, TVAL-3, D-AMP are implemented in Matlab. ReconNet, ISTA-Net, LAPRAN are implemented in Python with Pytorch\cite{paszke2019pytorch}. CSGM, CSGAN, LDAMP are implemented in Python with Tensorflow\cite{abadi2016tensorflow}. 

We provide a unified interface to run all the methods. Specifically, the common parameters of all methods are listed as follows: 
\begin{enumerate}
\item dataset: the name of dataset to be used
\item input\_channel: number of channels training/testing images have
\item input\_width: width of training/testing images
\item input\_height: height of training/testing images
\item m: number of measurements/outputs of sensing matrix
\item n: number of inputs to sensing matrix
\end{enumerate}
Besides, there are method-specific parameters that are included in a container-like object called "specifics". In python, it is a dictionary with its keys as parameter names and its values as actual parameters. In Matlab, it is a structure array with its field names as parameter names, and its field values are the parameters.

We also provide the functionality to directly call certain methods from the main interface. The parameters of the main interface are listed below:
\begin{enumerate}
\item sensing: method of sensing
\item reconstruction: method of reconstruction
\item stage: training or testing(model-based methods do not have this parameter)
\item default: will use default parameters if it's true. Will override other parameters set manually.
\item dataset: same as method's corresponding parameter. 
\item input\_channel: same as method's corresponding parameter.
\item input\_width: same as method's corresponding parameter.
\item input\_height: same as method's corresponding parameter.
\item m: same as method's corresponding parameter.
\item n: same as method's corresponding parameter.
\item specifics: specific parameter settings of chosen reconstruction method. Will be passed to the actual method.
\end{enumerate}

Given an image to be sensed and reconstructed, for model-based methods, it can be directly reconstructed by calling the main function of specific methods. For data-driven methods, the networks of specific methods have to be trained first. We provide pre-trained networks of each method at five compression ratios(2,4,8,16,32) on six datasets. We also provide the functionality of training new networks on 
new datasets and compression ratios from scratch.

More details regarding how to use our code are listed in the main page of our Github repository(\url{https://github.com/PSCLab-ASU/OpenICS}).

\section{Benchmarks}
\subsection{Benchmark Design}
%1. datasets, 2. implementation details(how is the methods implemented, how much has been modified) 3. Evaluation metrics.
\textbf{Dataset.} We use six widely used datasets to evaluate all the methods in benchmark. They are MNIST\cite{mnist}, CIFAR10\cite{cifar10}, CIFAR10(grayscaled), CELEBA\cite{celeba}, Bigset, Bigset(grayscaled). Bigset stands for a manually composed dataset. It was initially used in \cite{bigset1,bigset2,bigset3} in the domain of single image super-resolution. Later it was used in LAPRAN\cite{lapran} for image compressive sensing. The training set of Bigset was composed of 91 images
from \cite{bigsettrain} and 200 images from the BSD\cite{bsd} dataset. The 291 images are augmented (rotation and flip) and cut into 228688
patches as training samples. The testing set of Bigset consists of image patches from Set5\cite{set5} and Set14\cite{set14} with same patch size. For MNIST, CIFAR10, CIFAR10(gray), the image size of samples is 32x32. For CELEBA, Bigset(gray) and Bigset, the image size of samples is 64x64. 

\textbf{Compression ratios.} We take five different compression ratios to evaluate each method: 2,4,8,16,32. The compression is always performed channel-wise, i.e., for colored images(with RGB color channels), we perform the compression over each channel seperately. The measurements of all three channels are then grouped together for subsequent reconstruction. The training procedure of each data-driven method is almost the same as the original training guideline provided by the original authors. The discrepancy 
is detailed in our github repository. 

\textbf{Metrics.} We evaluate all the methods from two aspects: reconstruction accuracy and reconstruction speed. The reconstruction accuracy is quantified with two metrics: PSNR(0-48) and SSIM(0-1) between reconstructed images and original images in the testing set. Higher values indicate higher accuracy. For each experiment we conduct, the reported results are the averaged values of both metrics over all the samples of the corresponding testing set. The reconstruction speed is quantified with the number of images reconstructed per second. This value is averaged over all the samples in the testing set as well. 

\textbf{Benchmark calculation.} After obtaining the all raw benchmark results of each method(total of $6 \text{\ datasets} \times 5 \text{\ compression ratios} \times 3 \text{\ metrics} = 90 \text{\ raw results}$), we use the following equation to calculate the final benchmark score: 
\begin{equation}
    score=\sum_{i=1}^{90}w_{dataset}*w_{cr}*w_{metric}*\bar{v}_i
\end{equation}

 $\bar{v}_i$ is the ith normalized raw experiment result. Due to the different value ranges of each metric, we have to normalize the raw values to 0-100 range to avoid the dominance of one metric over the others. The function used to normalize PSNR values is $\bar{v}=10^{\frac{v}{48}-1}*100$. The function used to normalize SSIM values is $\bar{v}=10^{v-1}*100$. The function used to normalize reconstruction speed values is $\bar{v}=\frac{100}{1+1/log(1+v)}$. 
 
 $w_{dataset}$ is the weight of the corresponding dataset of $v_i$. Different weights are assigned to different datasets according to their relative complexity in reconstruction compared with other datasets. The relative complexity is determined based on the results reported in literatures\cite{l1,tval3,nlrcs,damp,reconnet,istanet,ldamp,lapran,csgm,csgan} in the domain of image compressive sensing.
 
 $w_{cr}$ is the weight of the corresponding compression ratio of $v_i$. Since images compressed at higher compression ratios are more difficult to reconstruct than images compressed at lower compression ratios, we assign higher weights to higher compression ratios. 
 
 $w_{metric}$ is the weight of corresponding metric of $v_i$. we assign different weights to different metrics as $\text{PSNR}:\text{SSIM}:\text{Speed}=1:1:2$. As such, there is no bias between reconstruction accuracy and reconstruction speed. One can specify own weights to different metrics to make the score reflects one's own preferences. 

The actual weights are listed in table~\ref{wdataset},table~\ref{wcr} and table~\ref{wmetric}.
\begin{table}[]
\centering
\begin{tabular}{|l|l|}
\hline
\multicolumn{1}{|c|}{\textbf{Dataset}} & \multicolumn{1}{c|}{\textbf{Weight}} \\ \hline
MNIST                                  & 1/21                                 \\ \hline
CelebA                                 & 4/21                                 \\ \hline
CIFAR10                                & 3/21                                 \\ \hline
CIFAR10 Gray                           & 2/21                                 \\ \hline
Bigset                                 & 6/21                                 \\ \hline
Bigset Gray                            & 5/21                                 \\ \hline
\end{tabular}
\vspace{4mm}
\caption{Weights of Datasets}
\label{wdataset}
\end{table}
\begin{table}[]
\centering
\begin{tabular}{|l|l|}
\hline
\multicolumn{1}{|c|}{\textbf{Compression ratio}} & \multicolumn{1}{c|}{\textbf{Weight}} \\ \hline
2                                    & 1/31                                 \\ \hline
4                                    & 2/31                                 \\ \hline
8                                    & 4/31                                 \\ \hline
16                                   & 8/31                                 \\ \hline
32                                   & 16/31                                \\ \hline
\end{tabular}
\vspace{4mm}
\caption{Weights of compression ratios}
\label{wcr}
\end{table}
\begin{table}[]
\centering
\begin{tabular}{|l|l|}
\hline
\multicolumn{1}{|c|}{\textbf{Metric}} & \multicolumn{1}{c|}{\textbf{Weight}} \\ \hline
PSNR                                  & 1/4                                  \\ \hline
SSIM                                  & 1/4                                  \\ \hline
Reconstruction speed                   & 1/2                                  \\ \hline
\end{tabular}
\vspace{4mm}
\caption{Weights of metrics}
\label{wmetric}
\end{table}
\subsection{Benchmark Results}
\begin{figure}
    \centering
    \includegraphics[width=\linewidth]{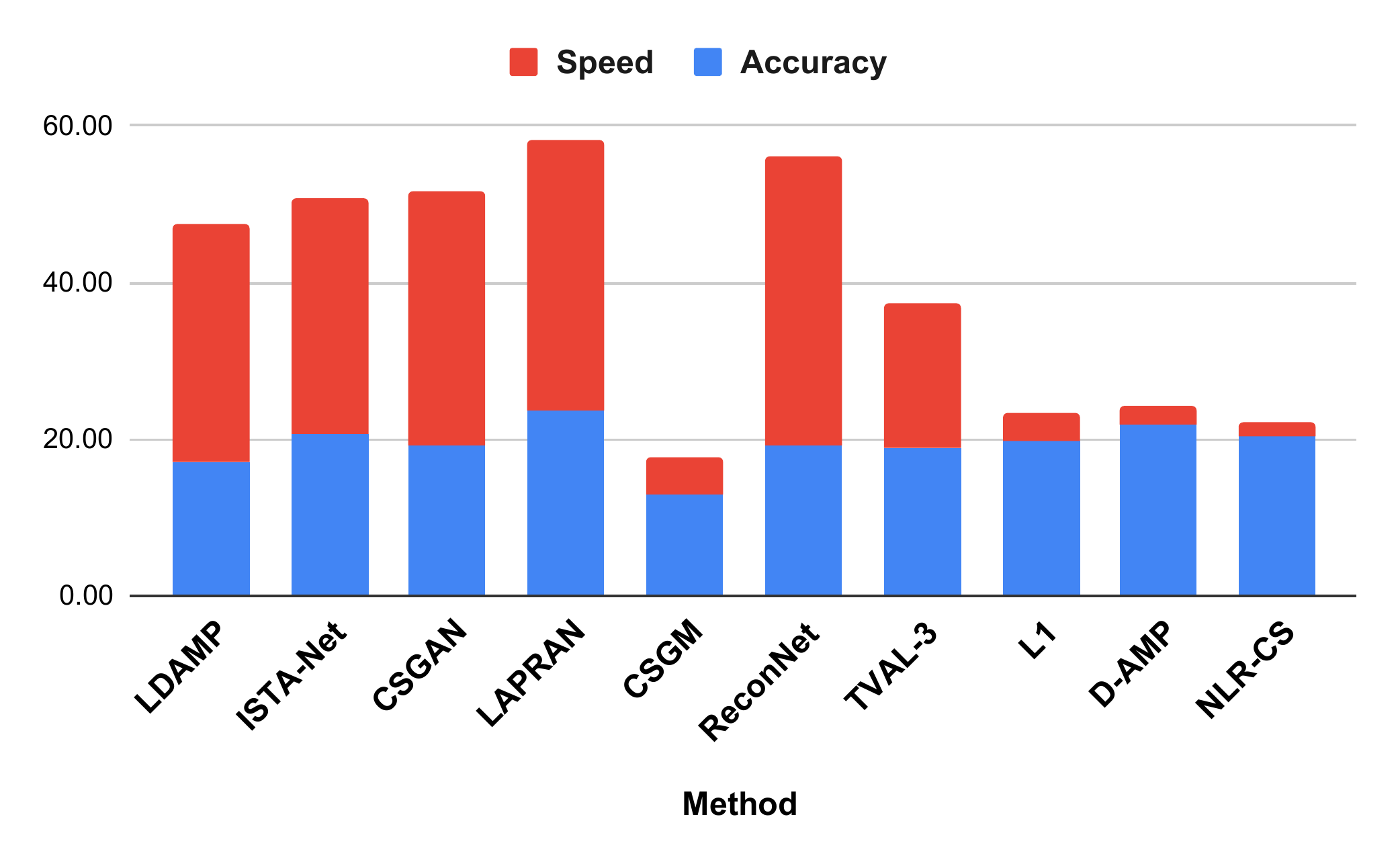}
    \caption{The benchamark scores}
    \label{histgram}
\end{figure}

The raw benchmark results are listed in Table~\ref{ldampresults},\ref{istanetresults},\ref{csganresults},\ref{lapranresults},\ref{csgmresults},\ref{reconnetresults},\ref{tval3results},\ref{l1results},\ref{dampresults} and \ref{nlrcsresults} in the appendix. The benchmark score of each method is shown in Table~\ref{benchmark} and Fig~\ref{histgram}.

\begin{table}[]
\centering
\begin{tabular}{|l|r|r|r|}
\hline
\multicolumn{1}{|c|}{Method} & \multicolumn{1}{c|}{Speed} & \multicolumn{1}{c|}{Accuracy} & \multicolumn{1}{c|}{Score} \\ \hline
LDAMP                        & 30.25                      & 17.21                         & 47.46                      \\ \hline
ISTA-Net                     & 30.02                      & 20.69                         & 50.71                      \\ \hline
CSGAN                        & 32.58                      & 19.03                         & 51.61                      \\ \hline
LAPRAN                       & 34.69                      & 23.60                         & 58.30                      \\ \hline
CSGM                         & 4.75                       & 13.07                         & 17.82                      \\ \hline
ReconNet                     & 37.00                      & 19.15                         & 56.15                      \\ \hline
TVAL-3                       & 18.43                      & 18.92                         & 37.35                      \\ \hline
L1                           & 3.78                       & 19.69                         & 23.46                      \\ \hline
D-AMP                        & 2.35                       & 21.83                         & 24.19                      \\ \hline
NLR-CS                       & 1.69                       & 20.35                         & 22.04                      \\ \hline
\end{tabular}
\vspace{4mm}
\caption{The benchmark scores}
\label{benchmark}
\end{table}
LAPRAN has the highest benchmark score due to its prominent performance in accuracy and speed. LDAMP has the highest performance in accuracy but bad performance in speed due to its heavyweight design of network structure(more than 200 neural layers). ReconNet has the highest performance in reconstruction speed due to its lightweight design in structure(only seven layers), but its performance in accuracy is limited as well. In general, model-based methods have lower performance on both accuracy and speed than data-driven methods due to their static, pre-defined signal prior and iterative running process. CSGM is a special case in data-driven methods. The unsatisfying performance in accuracy is due to the GAN model it uses, which is DCGAN\cite{dcgan} proposed in 2015. Over the past few years, there have been more successful GAN models proposed, such as StyleGAN\cite{stylegan} that has much higher performance in modeling signals from data, which may improve the performance of CSGM if it is used. For all the model-based methods, NLR-CS and D-AMP have higher performance in reconstruction accuracy but lower performance in reconstruction speed compared with the other two methods.

To conclude, in general, data-driven methods achieve the highest performance in terms of accuracy and performance. With enough training data and hardware platforms that have sufficient computation capacity, one should always choose end-to-end data-driven methods. If there is no sufficient data, one should choose model-based methods that have the highest reconstruction accuracy. If the reconstruction speed is a critical factor to consider as well, TVAL-3 has a significantly higher reconstruction speed than other model-based methods and comparable reconstruction accuracy to other methods.

\section*{Acknowledgment}
This work is supported by the Research Experiences for Undergraduates (REU) funding of an NSF grant (IIS/CPS-1652038) and the Fulton Undergraduate Research Initiative (FURI) program at Arizona State University. Part of the NVIDIA GPUs used for this work was donated by NVIDIA Corporation. The CPU servers used for this work were donated by Intel Corporation. 
\bibliographystyle{IEEEtran}
\bibliography{ref}
\section{Appendix}

\begin{table*}[]
\centering
\begin{tabular}{|c|c|r|r|r|}
\hline
              & Compression ratio & \multicolumn{1}{c|}{PSNR}       & \multicolumn{1}{c|}{SSIM}      & \multicolumn{1}{c|}{Reconstruction per second} \\ \hline
MNIST         & 2                 & 47.9998                         & 0.9997                         & 27.5482                                        \\ \hline
MNIST         & 4                 & 47.5808                         & 0.9994                         & 34.6021                                        \\ \hline
MNIST         & 8                 & 39.8992                         & 0.9965                         & 34.9650                                        \\ \hline
MNIST         & 16                & 30.8985                         & 0.9751                         & 34.7222                                        \\ \hline
MNIST         & 32                & 16.2898                         & 0.5901                         & 34.3643                                        \\ \hline
CelebA        & 2                 & 41.1183                         & 0.9911                         & 23.4192                                        \\ \hline
CelebA        & 4                 & 32.0614                         & 0.9459                         & 27.5482                                        \\ \hline
CelebA        & 8                 & 27.5465                         & 0.8723                         & 30.9598                                        \\ \hline
CelebA        & 16                & 23.592                          & 0.7415                         & 33.5570                                        \\ \hline
CelebA        & 32                & 17.5607                         & 0.3732                         & 32.5733                                        \\ \hline
CIFAR10       & 2                 & 35.1695                         & 0.9748                         & 35.2113                                        \\ \hline
CIFAR10       & 4                 & 28.5017                         & 0.9052                         & 35.4610                                        \\ \hline
CIFAR10       & 8                 & 22.7743                         & 0.7285                         & 33.5570                                        \\ \hline
CIFAR10       & 16                & 18.8613                         & 0.498                          & 34.6021                                        \\ \hline
CIFAR10       & 32                & 13.7782                         & 0.1406                         & 34.1297                                        \\ \hline
CIFAR10(Gray) & 2                 & 34.3648                         & 0.9713                         & 32.5733                                        \\ \hline
CIFAR10(Gray) & 4                 & 27.8835                         & 0.8945                         & 33.0033                                        \\ \hline
CIFAR10(Gray) & 8                 & 23.2726                         & 0.7499                         & 34.4828                                        \\ \hline
CIFAR10(Gray) & 16                & 16.6404                         & 0.3471                         & 35.2113                                        \\ \hline
CIFAR10(Gray) & 32                & 12.2314                         & 0.0701                         & 35.0877                                        \\ \hline
Bigset        & 2                 & 38.4191                         & 0.9432                         & 23.9234                                        \\ \hline
Bigset        & 4                 & 34.3927                         & 0.8936                         & 29.1545                                        \\ \hline
Bigset        & 8                 & 30.9195                         & 0.8096                         & 29.2398                                        \\ \hline
Bigset        & 16                & 27.608                          & 0.7055                         & 34.3643                                        \\ \hline
Bigset        & 32                & 17.3593                         & 0.1853                         & 33.2226                                        \\ \hline
Bigset(Gray)  & 2                 & 38.5702                         & 0.9508                         & 22.8311                                        \\ \hline
Bigset(Gray)  & 4                 & 34.5885                         & 0.8956                         & 27.3224                                        \\ \hline
Bigset(Gray)  & 8                 & 31.0164                         & 0.808                          & 30.1205                                        \\ \hline
Bigset(Gray)  & 16                & 28.3581                         & 0.7207                         & 33.4448                                        \\ \hline
Bigset(Gray)  & 32                & 17.283                          & 0.178                          & 35.5872                                        \\ \hline
\end{tabular}
\vspace{4mm}
\caption{Benchmark results of LDAMP}
\label{ldampresults}
\end{table*}

\begin{table*}[]
\centering
\begin{tabular}{|c|c|r|r|r|}
\hline
Dataset       & Compression ratio & \multicolumn{1}{c|}{PSNR} & \multicolumn{1}{c|}{SSIM} & \multicolumn{1}{c|}{Reconstruction per second} \\ \hline
MNIST         & 2                 & 47.99                     & 0.9999                    & 93.4579                                        \\ \hline
MNIST         & 4                 & 44.27                     & 0.9988                    & 55.2486                                        \\ \hline
MNIST         & 8                 & 35.12                     & 0.9907                    & 75.7576                                        \\ \hline
MNIST         & 16                & 27.31                     & 0.9532                    & 75.7576                                        \\ \hline
MNIST         & 32                & 19.76                     & 0.7747                    & 82.6446                                        \\ \hline
CelebA        & 2                 & 37.43                     & 0.9798                    & 11.9190                                        \\ \hline
CelebA        & 4                 & 31.14                     & 0.9297                    & 14.3062                                        \\ \hline
CelebA        & 8                 & 27.07                     & 0.8499                    & 4.8170                                         \\ \hline
CelebA        & 16                & 23.77                     & 0.7406                    & 7.3475                                         \\ \hline
CelebA        & 32                & 21.13                     & 0.6295                    & 17.5439                                        \\ \hline
CIFAR10       & 2                 & 34.12                     & 0.9703                    & 30.1205                                        \\ \hline
CIFAR10       & 4                 & 27.66                     & 0.8932                    & 33.2226                                        \\ \hline
CIFAR10       & 8                 & 23.41                     & 0.7632                    & 31.4465                                        \\ \hline
CIFAR10       & 16                & 20.25                     & 0.5979                    & 25.9067                                        \\ \hline
CIFAR10       & 32                & 17.95                     & 0.435                     & 26.3852                                        \\ \hline
CIFAR10(Gray) & 2                 & 33.63                     & 0.9679                    & 138.8889                                       \\ \hline
CIFAR10(Gray) & 4                 & 27.46                     & 0.8886                    & 68.0272                                        \\ \hline
CIFAR10(Gray) & 8                 & 23.15                     & 0.7501                    & 75.1880                                        \\ \hline
CIFAR10(Gray) & 16                & 20.25                     & 0.5911                    & 81.3008                                        \\ \hline
CIFAR10(Gray) & 32                & 18.13                     & 0.4406                    & 86.2069                                        \\ \hline
Bigset        & 2                 & 37.28                     & 0.9393                    & 15.8479                                        \\ \hline
Bigset        & 4                 & 33                        & 0.8686                    & 19.5313                                        \\ \hline
Bigset        & 8                 & 29.88                     & 0.7823                    & 22.0751                                        \\ \hline
Bigset        & 16                & 27.03                     & 0.682                     & 25.6410                                        \\ \hline
Bigset        & 32                & 24.67                     & 0.5864                    & 29.7619                                        \\ \hline
Bigset(Gray)  & 2                 & 38.49                     & 0.95                      & 57.8035                                        \\ \hline
Bigset(Gray)  & 4                 & 34.06                     & 0.8874                    & 76.9231                                        \\ \hline
Bigset(Gray)  & 8                 & 30.69                     & 0.8007                    & 82.6446                                        \\ \hline
Bigset(Gray)  & 16                & 27.66                     & 0.7035                    & 74.0741                                        \\ \hline
Bigset(Gray)  & 32                & 25.16                     & 0.6074                    & 65.3595                                        \\ \hline
\end{tabular}
\vspace{4mm}
\caption{Benchmark results of ISTA-Net}
\label{istanetresults}
\end{table*}

\begin{table*}[]
\centering
\begin{tabular}{|c|c|r|r|r|}
\hline
Dataset       & Compression ratio & \multicolumn{1}{c|}{PSNR} & \multicolumn{1}{c|}{SSIM} & \multicolumn{1}{c|}{Reconstruction per second} \\ \hline
MNIST         & 2                 & 29.2126                   & 0.9640                    & 118.4133                                       \\ \hline
MNIST         & 4                 & 30.5120                   & 0.9563                    & 120.9862                                       \\ \hline
MNIST         & 8                 & 29.1593                   & 0.9558                    & 123.8261                                       \\ \hline
MNIST         & 16                & 25.2426                   & 0.9275                    & 122.9381                                       \\ \hline
MNIST         & 32                & 21.3493                   & 0.8455                    & 121.1784                                       \\ \hline
CelebA        & 2                 & 20.2156                   & 0.8108                    & 71.2754                                        \\ \hline
CelebA        & 4                 & 17.7137                   & 0.6881                    & 73.5295                                        \\ \hline
CelebA        & 8                 & 18.0141                   & 0.7477                    & 71.7567                                        \\ \hline
CelebA        & 16                & 21.0598                   & 0.8439                    & 73.0777                                        \\ \hline
CelebA        & 32                & 21.5447                   & 0.8538                    & 71.7036                                        \\ \hline
CIFAR10       & 2                 & 19.8350                   & 0.8110                    & 70.4578                                        \\ \hline
CIFAR10       & 4                 & 21.6027                   & 0.8646                    & 69.8564                                        \\ \hline
CIFAR10       & 8                 & 21.7896                   & 0.8692                    & 71.6831                                        \\ \hline
CIFAR10       & 16                & 21.2262                   & 0.8506                    & 72.7173                                        \\ \hline
CIFAR10       & 32                & 19.7638                   & 0.7946                    & 71.1875                                        \\ \hline
CIFAR10(Gray) & 2                 & 23.4110                   & 0.7581                    & 75.6761                                        \\ \hline
CIFAR10(Gray) & 4                 & 22.8502                   & 0.7318                    & 75.1091                                        \\ \hline
CIFAR10(Gray) & 8                 & 21.7888                   & 0.6794                    & 74.7101                                        \\ \hline
CIFAR10(Gray) & 16                & 19.7298                   & 0.5493                    & 74.7030                                        \\ \hline
CIFAR10(Gray) & 32                & 18.1149                   & 0.4288                    & 74.9074                                        \\ \hline
Bigset        & 2                 & 18.1686                   & 0.5370                    & 72.5171                                        \\ \hline
Bigset        & 4                 & 16.3798                   & 0.3874                    & 80.3709                                        \\ \hline
Bigset        & 8                 & 17.6498                   & 0.4811                    & 72.8396                                        \\ \hline
Bigset        & 16                & 21.4860                   & 0.6384                    & 72.8716                                        \\ \hline
Bigset        & 32                & 21.0166                   & 0.6326                    & 66.1600                                        \\ \hline
Bigset(Gray)  & 2                 & 18.2207                   & 0.2187                    & 70.1563                                        \\ \hline
Bigset(Gray)  & 4                 & 21.5229                   & 0.3874                    & 68.3210                                        \\ \hline
Bigset(Gray)  & 8                 & 23.2503                   & 0.5378                    & 71.6273                                        \\ \hline
Bigset(Gray)  & 16                & 23.5321                   & 0.5401                    & 72.4786                                        \\ \hline
Bigset(Gray)  & 32                & 22.8348                   & 0.5089                    & 71.2181                                        \\ \hline
\end{tabular}
\vspace{4mm}
\caption{Benchmark results of CSGAN}
\label{csganresults}
\end{table*}

\begin{table*}[]
\centering
\begin{tabular}{|c|c|r|r|r|}
\hline
Dataset       & Compression ratio & \multicolumn{1}{c|}{PSNR} & \multicolumn{1}{c|}{SSIM} & \multicolumn{1}{c|}{Reconstruction per second} \\ \hline
MNIST         & 2                 & 32.0483                   & 0.9156                    & 223.1187                                       \\ \hline
MNIST         & 4                 & 32.1388                   & 0.9869                    & 221.9279                                       \\ \hline
MNIST         & 8                 & 26.5582                   & 0.954                     & 222.8450                                       \\ \hline
MNIST         & 16                & 23.3674                   & 0.8691                    & 229.3073                                       \\ \hline
MNIST         & 32                & 19.7423                   & 0.7701                    & 219.0102                                       \\ \hline
CelebA        & 2                 & 29.4438                   & 0.975                     & 137.9121                                       \\ \hline
CelebA        & 4                 & 33.2433                   & 0.9888                    & 150.3084                                       \\ \hline
CelebA        & 8                 & 28.9183                   & 0.9698                    & 146.9329                                       \\ \hline
CelebA        & 16                & 26.2793                   & 0.9483                    & 151.5757                                       \\ \hline
CelebA        & 32                & 23.1517                   & 0.8969                    & 151.6358                                       \\ \hline
CIFAR10       & 2                 & 31.6029                   & 0.9862                    & 174.7553                                       \\ \hline
CIFAR10       & 4                 & 27.808                    & 0.9674                    & 225.8276                                       \\ \hline
CIFAR10       & 8                 & 25.4534                   & 0.9428                    & 227.8823                                       \\ \hline
CIFAR10       & 16                & 22.262                    & 0.8844                    & 230.9234                                       \\ \hline
CIFAR10       & 32                & 19.852                    & 0.7874                    & 214.7841                                       \\ \hline
CIFAR10(Gray) & 2                 & 25.7358                   & 0.8669                    & 141.5323                                       \\ \hline
CIFAR10(Gray) & 4                 & 23.6409                   & 0.7804                    & 219.8296                                       \\ \hline
CIFAR10(Gray) & 8                 & 21.621                    & 0.6675                    & 229.1173                                       \\ \hline
CIFAR10(Gray) & 16                & 19.6054                   & 0.5293                    & 223.8697                                       \\ \hline
CIFAR10(Gray) & 32                & 18.0317                   & 0.3887                    & 232.5259                                       \\ \hline
Bigset        & 2                 & 30.7518                   & 0.9314                    & 30.5301                                        \\ \hline
Bigset        & 4                 & 30.2502                   & 0.9214                    & 34.1619                                        \\ \hline
Bigset        & 8                 & 28.3798                   & 0.881                     & 36.4556                                        \\ \hline
Bigset        & 16                & 27.3575                   & 0.8539                    & 34.5518                                        \\ \hline
Bigset        & 32                & 24.035                    & 0.7489                    & 41.8374                                        \\ \hline
Bigset(Gray)  & 2                 & 30.5937                   & 0.8376                    & 35.3011                                        \\ \hline
Bigset(Gray)  & 4                 & 30.2819                   & 0.8125                    & 41.4439                                        \\ \hline
Bigset(Gray)  & 8                 & 27.0031                   & 0.7037                    & 32.2543                                        \\ \hline
Bigset(Gray)  & 16                & 25.1241                   & 0.6216                    & 41.1410                                        \\ \hline
Bigset(Gray)  & 32                & 23.6821                   & 0.5611                    & 44.2111                                        \\ \hline
\end{tabular}
\vspace{4mm}
\caption{Benchmark results of LAPRAN}
\label{lapranresults}
\end{table*}

\begin{table*}[]
\centering
\begin{tabular}{|c|c|r|r|r|}
\hline
Dataset       & Compression ratio & \multicolumn{1}{c|}{PSNR} & \multicolumn{1}{c|}{SSIM} & \multicolumn{1}{c|}{Reconstruction per second} \\ \hline
MNIST         & 2                 & 22.6164                   & 0.8978                    & 2.0329                                         \\ \hline
MNIST         & 4                 & 22.4511                   & 0.8923                    & 2.3868                                         \\ \hline
MNIST         & 8                 & 21.9578                   & 0.8742                    & 2.6208                                         \\ \hline
MNIST         & 16                & 20.6622                   & 0.8215                    & 2.6924                                         \\ \hline
MNIST         & 32                & 17.7698                   & 0.6985                    & 2.8841                                         \\ \hline
CelebA        & 2                 & 21.0459                   & 0.6081                    & 0.0178                                         \\ \hline
CelebA        & 4                 & 20.9366                   & 0.6034                    & 0.0405                                         \\ \hline
CelebA        & 8                 & 20.7178                   & 0.5938                    & 0.0421                                         \\ \hline
CelebA        & 16                & 20.2657                   & 0.5737                    & 0.0661                                         \\ \hline
CelebA        & 32                & 19.3262                   & 0.5306                    & 0.1634                                         \\ \hline
CIFAR10       & 2                 & 19.3796                   & 0.5837                    & 0.3100                                         \\ \hline
CIFAR10       & 4                 & 19.0948                   & 0.5679                    & 0.3939                                         \\ \hline
CIFAR10       & 8                 & 18.5491                   & 0.5369                    & 0.4501                                         \\ \hline
CIFAR10       & 16                & 17.4882                   & 0.4752                    & 0.4861                                         \\ \hline
CIFAR10       & 32                & 15.7465                   & 0.3754                    & 0.5013                                         \\ \hline
CIFAR10(Gray) & 2                 & 18.8775                   & 0.5630                    & 0.4896                                         \\ \hline
CIFAR10(Gray) & 4                 & 18.1176                   & 0.5190                    & 0.5027                                         \\ \hline
CIFAR10(Gray) & 8                 & 16.7601                   & 0.4420                    & 0.5116                                         \\ \hline
CIFAR10(Gray) & 16                & 14.7029                   & 0.3264                    & 0.5160                                         \\ \hline
CIFAR10(Gray) & 32                & 12.4420                   & 0.2042                    & 0.5175                                         \\ \hline
Bigset        & 2                 & 20.5856                   & 0.4296                    & 0.0294                                         \\ \hline
Bigset        & 4                 & 20.5270                   & 0.4275                    & 0.0546                                         \\ \hline
Bigset        & 8                 & 20.3866                   & 0.4207                    & 0.0773                                         \\ \hline
Bigset        & 16                & 20.0988                   & 0.4054                    & 0.1022                                         \\ \hline
Bigset        & 32                & 19.5600                   & 0.3787                    & 0.1551                                         \\ \hline
Bigset(Gray)  & 2                 & 20.5140                   & 0.4439                    & 0.1111                                         \\ \hline
Bigset(Gray)  & 4                 & 20.2344                   & 0.4295                    & 0.1688                                         \\ \hline
Bigset(Gray)  & 8                 & 19.7265                   & 0.4028                    & 0.2307                                         \\ \hline
Bigset(Gray)  & 16                & 18.6832                   & 0.3527                    & 0.2580                                         \\ \hline
Bigset(Gray)  & 32                & 16.8286                   & 0.2712                    & 0.2697                                         \\ \hline
\end{tabular}
\vspace{4mm}
\caption{Benchmark results of CSGM}
\label{csgmresults}
\end{table*}

\begin{table*}[]
\centering
\begin{tabular}{|c|c|r|r|r|}
\hline
Dataset       & Compression ratio & \multicolumn{1}{c|}{PSNR} & \multicolumn{1}{c|}{SSIM} & \multicolumn{1}{c|}{Reconstruction per second} \\ \hline
MNIST         & 2                 & 38.471                    & 0.985                     & 723.5890                                       \\ \hline
MNIST         & 4                 & 32.228                    & 0.984                     & 874.8906                                       \\ \hline
MNIST         & 8                 & 27.558                    & 0.933                     & 848.8964                                       \\ \hline
MNIST         & 16                & 23.860                    & 0.914                     & 688.7052                                       \\ \hline
MNIST         & 32                & 20.259                    & 0.821                     & 712.2507                                       \\ \hline
CelebA        & 2                 & 33.390                    & 0.954                     & 604.5949                                       \\ \hline
CelebA        & 4                 & 28.623                    & 0.889                     & 621.8905                                       \\ \hline
CelebA        & 8                 & 25.595                    & 0.812                     & 611.2469                                       \\ \hline
CelebA        & 16                & 23.115                    & 0.722                     & 791.1392                                       \\ \hline
CelebA        & 32                & 21.065                    & 0.634                     & 623.0530                                       \\ \hline
CIFAR10       & 2                 & 30.494                    & 0.945                     & 552.1811                                       \\ \hline
CIFAR10       & 4                 & 25.468                    & 0.847                     & 744.6016                                       \\ \hline
CIFAR10       & 8                 & 22.274                    & 0.719                     & 807.1025                                       \\ \hline
CIFAR10       & 16                & 19.836                    & 0.570                     & 708.2153                                       \\ \hline
CIFAR10       & 32                & 17.957                    & 0.430                     & 777.6050                                       \\ \hline
CIFAR10(Gray) & 2                 & 30.662                    & 0.946                     & 683.0601                                       \\ \hline
CIFAR10(Gray) & 4                 & 25.466                    & 0.842                     & 689.6552                                       \\ \hline
CIFAR10(Gray) & 8                 & 22.584                    & 0.723                     & 802.5682                                       \\ \hline
CIFAR10(Gray) & 16                & 20.080                    & 0.571                     & 736.9197                                       \\ \hline
CIFAR10(Gray) & 32                & 18.264                    & 0.433                     & 803.8585                                       \\ \hline
Bigset        & 2                 & 32.544                    & 0.873                     & 798.7220                                       \\ \hline
Bigset        & 4                 & 28.862                    & 0.782                     & 805.1530                                       \\ \hline
Bigset        & 8                 & 26.872                    & 0.705                     & 772.7975                                       \\ \hline
Bigset        & 16                & 24.945                    & 0.618                     & 655.3080                                       \\ \hline
Bigset        & 32                & 23.377                    & 0.556                     & 661.3757                                       \\ \hline
Bigset(Gray)  & 2                 & 34.356                    & 0.911                     & 566.5722                                       \\ \hline
Bigset(Gray)  & 4                 & 31.002                    & 0.832                     & 796.8127                                       \\ \hline
Bigset(Gray)  & 8                 & 28.232                    & 0.742                     & 712.2507                                       \\ \hline
Bigset(Gray)  & 16                & 26.024                    & 0.653                     & 662.6905                                       \\ \hline
Bigset(Gray)  & 32                & 24.000                    & 0.569                     & 696.3788                                       \\ \hline
\end{tabular}
\vspace{4mm}
\caption{Benchmark results of ReconNet}
\label{reconnetresults}
\end{table*}

\begin{table*}[]
\centering
\begin{tabular}{|c|c|r|r|r|}
\hline
Dataset       & Compression ratio & \multicolumn{1}{c|}{PSNR} & \multicolumn{1}{c|}{SSIM} & \multicolumn{1}{c|}{Reconstruction per second} \\ \hline
MNIST         & 2                 & 47.995                    & 1                         & 16.3934                                        \\ \hline
MNIST         & 4                 & 33.233                    & 0.879                     & 12.3457                                        \\ \hline
MNIST         & 8                 & 20.587                    & 0.542                     & 12.3457                                        \\ \hline
MNIST         & 16                & 15.291                    & 0.299                     & 15.8730                                        \\ \hline
MNIST         & 32                & 13.076                    & 0.163                     & 16.1290                                        \\ \hline
CelebA        & 2                 & 32.335                    & 0.959                     & 0.7283                                         \\ \hline
CelebA        & 4                 & 26.592                    & 0.889                     & 1.2516                                         \\ \hline
CelebA        & 8                 & 22.863                    & 0.801                     & 1.2563                                         \\ \hline
CelebA        & 16                & 19.919                    & 0.703                     & 1.4104                                         \\ \hline
CelebA        & 32                & 17.345                    & 0.599                     & 1.4025                                         \\ \hline
CIFAR10       & 2                 & 29.584                    & 0.936                     & 4.6948                                         \\ \hline
CIFAR10       & 4                 & 24.001                    & 0.822                     & 4.6083                                         \\ \hline
CIFAR10       & 8                 & 20.621                    & 0.69                      & 4.8077                                         \\ \hline
CIFAR10       & 16                & 18.286                    & 0.573                     & 5.2356                                         \\ \hline
CIFAR10       & 32                & 16.401                    & 0.476                     & 5.3476                                         \\ \hline
CIFAR10(Gray) & 2                 & 29.766                    & 0.9                       & 14.4928                                        \\ \hline
CIFAR10(Gray) & 4                 & 24.189                    & 0.742                     & 13.1579                                        \\ \hline
CIFAR10(Gray) & 8                 & 20.778                    & 0.577                     & 13.6986                                        \\ \hline
CIFAR10(Gray) & 16                & 18.343                    & 0.446                     & 15.3846                                        \\ \hline
CIFAR10(Gray) & 32                & 16.661                    & 0.362                     & 16.1290                                        \\ \hline
Bigset        & 2                 & 35.83                     & 0.96                      & 0.7962                                         \\ \hline
Bigset        & 4                 & 31.084                    & 0.905                     & 1.3106                                         \\ \hline
Bigset        & 8                 & 27.632                    & 0.845                     & 1.3298                                         \\ \hline
Bigset        & 16                & 24.754                    & 0.786                     & 1.4286                                         \\ \hline
Bigset        & 32                & 22.109                    & 0.729                     & 1.4225                                         \\ \hline
Bigset(Gray)  & 2                 & 36.154                    & 0.915                     & 1.6447                                         \\ \hline
Bigset(Gray)  & 4                 & 31.368                    & 0.814                     & 3.5971                                         \\ \hline
Bigset(Gray)  & 8                 & 27.871                    & 0.711                     & 3.9063                                         \\ \hline
Bigset(Gray)  & 16                & 24.954                    & 0.62                      & 4.2735                                         \\ \hline
Bigset(Gray)  & 32                & 22.054                    & 0.545                     & 4.0984                                         \\ \hline
\end{tabular}
\vspace{4mm}
\caption{Benchmark results of TVAL-3}
\label{tval3results}
\end{table*}

\begin{table*}[]
\centering
\begin{tabular}{|c|c|r|r|r|}
\hline
Dataset       & Compression ratio & \multicolumn{1}{c|}{PSNR} & \multicolumn{1}{c|}{SSIM} & \multicolumn{1}{c|}{Reconstruction per second} \\ \hline
MNIST         & 2                 & 47.911                    & 0.999                     & 0.8569                                         \\ \hline
MNIST         & 4                 & 31.01                     & 0.803                     & 0.8244                                         \\ \hline
MNIST         & 8                 & 19.881                    & 0.489                     & 0.8244                                         \\ \hline
MNIST         & 16                & 14.158                    & 0.225                     & 1.0132                                         \\ \hline
MNIST         & 32                & 11.762                    & 0.1                       & 1.1779                                         \\ \hline
CelebA        & 2                 & 32.578                    & 0.961                     & 0.0307                                         \\ \hline
CelebA        & 4                 & 27.003                    & 0.897                     & 0.0487                                         \\ \hline
CelebA        & 8                 & 23.359                    & 0.816                     & 0.0516                                         \\ \hline
CelebA        & 16                & 20.721                    & 0.731                     & 0.0531                                         \\ \hline
CelebA        & 32                & 18.235                    & 0.633                     & 0.0471                                         \\ \hline
CIFAR10       & 2                 & 30.078                    & 0.943                     & 0.2993                                         \\ \hline
CIFAR10       & 4                 & 24.486                    & 0.837                     & 0.3171                                         \\ \hline
CIFAR10       & 8                 & 21.115                    & 0.709                     & 0.2793                                         \\ \hline
CIFAR10       & 16                & 18.697                    & 0.584                     & 0.3167                                         \\ \hline
CIFAR10       & 32                & 16.746                    & 0.476                     & 0.3738                                         \\ \hline
CIFAR10(Gray) & 2                 & 30.231                    & 0.91                      & 0.7868                                         \\ \hline
CIFAR10(Gray) & 4                 & 24.619                    & 0.758                     & 1.0030                                         \\ \hline
CIFAR10(Gray) & 8                 & 21.328                    & 0.596                     & 0.9891                                         \\ \hline
CIFAR10(Gray) & 16                & 18.822                    & 0.452                     & 1.0194                                         \\ \hline
CIFAR10(Gray) & 32                & 17.046                    & 0.344                     & 1.1173                                         \\ \hline
Bigset        & 2                 & 36.084                    & 0.961                     & 0.0351                                         \\ \hline
Bigset        & 4                 & 31.6                      & 0.91                      & 0.0522                                         \\ \hline
Bigset        & 8                 & 28.475                    & 0.855                     & 0.0554                                         \\ \hline
Bigset        & 16                & 25.951                    & 0.802                     & 0.0568                                         \\ \hline
Bigset        & 32                & 23.821                    & 0.754                     & 0.0599                                         \\ \hline
Bigset(Gray)  & 2                 & 36.425                    & 0.919                     & 0.0862                                         \\ \hline
Bigset(Gray)  & 4                 & 31.881                    & 0.823                     & 0.1557                                         \\ \hline
Bigset(Gray)  & 8                 & 28.687                    & 0.727                     & 0.1642                                         \\ \hline
Bigset(Gray)  & 16                & 26.128                    & 0.64                      & 0.1740                                         \\ \hline
Bigset(Gray)  & 32                & 24.035                    & 0.57                      & 0.1671                                         \\ \hline
\end{tabular}
\vspace{4mm}
\caption{Benchmark results of L1}
\label{l1results}
\end{table*}

\begin{table*}[]
\centering
\begin{tabular}{|c|c|r|r|r|}
\hline
Dataset       & Compression ratio & \multicolumn{1}{c|}{PSNR} & \multicolumn{1}{c|}{SSIM} & \multicolumn{1}{c|}{Reconstruction per second} \\ \hline
MNIST         & 2                 & 43.45                     & 0.972                     & 0.3483                                         \\ \hline
MNIST         & 4                 & 31.787                    & 0.891                     & 0.3526                                         \\ \hline
MNIST         & 8                 & 22.582                    & 0.724                     & 0.3407                                         \\ \hline
MNIST         & 16                & 13.238                    & 0.322                     & 0.3274                                         \\ \hline
MNIST         & 32                & 6.53                      & 0.093                     & 0.3194                                         \\ \hline
CelebA        & 2                 & 47.126                    & 0.998                     & 0.0568                                         \\ \hline
CelebA        & 4                 & 37.495                    & 0.985                     & 0.0537                                         \\ \hline
CelebA        & 8                 & 31.095                    & 0.95                      & 0.0458                                         \\ \hline
CelebA        & 16                & 26.328                    & 0.877                     & 0.0450                                         \\ \hline
CelebA        & 32                & 21.517                    & 0.723                     & 0.0500                                         \\ \hline
CIFAR10       & 2                 & 40.401                    & 0.993                     & 0.2190                                         \\ \hline
CIFAR10       & 4                 & 31.595                    & 0.957                     & 0.2104                                         \\ \hline
CIFAR10       & 8                 & 25.851                    & 0.867                     & 0.1976                                         \\ \hline
CIFAR10       & 16                & 20.248                    & 0.645                     & 0.1882                                         \\ \hline
CIFAR10       & 32                & 8.307                     & 0.113                     & 0.1843                                         \\ \hline
CIFAR10(Gray) & 2                 & 31.682                    & 0.929                     & 0.3606                                         \\ \hline
CIFAR10(Gray) & 4                 & 25.616                    & 0.789                     & 0.3524                                         \\ \hline
CIFAR10(Gray) & 8                 & 19.996                    & 0.537                     & 0.3398                                         \\ \hline
CIFAR10(Gray) & 16                & 16.365                    & 0.342                     & 0.3418                                         \\ \hline
CIFAR10(Gray) & 32                & 8.016                     & 0.074                     & 0.3398                                         \\ \hline
Bigset        & 2                 & 42.51                     & 0.992                     & 0.0474                                         \\ \hline
Bigset        & 4                 & 37.565                    & 0.977                     & 0.0520                                         \\ \hline
Bigset        & 8                 & 33.95                     & 0.949                     & 0.0516                                         \\ \hline
Bigset        & 16                & 30.973                    & 0.908                     & 0.0480                                         \\ \hline
Bigset        & 32                & 27.3                      & 0.832                     & 0.0496                                         \\ \hline
Bigset(Gray)  & 2                 & 38.205                    & 0.94                      & 0.1224                                         \\ \hline
Bigset(Gray)  & 4                 & 34.138                    & 0.87                      & 0.1222                                         \\ \hline
Bigset(Gray)  & 8                 & 30.774                    & 0.786                     & 0.1216                                         \\ \hline
Bigset(Gray)  & 16                & 27.234                    & 0.67                      & 0.1221                                         \\ \hline
Bigset(Gray)  & 32                & 20.727                    & 0.435                     & 0.1216                                         \\ \hline
\end{tabular}
\vspace{4mm}
\caption{Benchmark results of D-AMP}
\label{dampresults}
\end{table*}

\begin{table*}[]
\centering
\begin{tabular}{|c|c|r|r|r|}
\hline
Dataset       & Compression ratio & \multicolumn{1}{c|}{PSNR} & \multicolumn{1}{c|}{SSIM} & \multicolumn{1}{c|}{Reconstruction per second} \\ \hline
MNIST         & 2                 & 40.062                    & 0.912                     & 0.3148                                         \\ \hline
MNIST         & 4                 & 29.452                    & 0.778                     & 0.3264                                         \\ \hline
MNIST         & 8                 & 18.787                    & 0.472                     & 0.3270                                         \\ \hline
MNIST         & 16                & 15.058                    & 0.312                     & 0.3247                                         \\ \hline
MNIST         & 32                & 12.052                    & 0.132                     & 0.3258                                         \\ \hline
CelebA        & 2                 & 35.932                    & 0.978                     & 0.0238                                         \\ \hline
CelebA        & 4                 & 29.288                    & 0.93                      & 0.0233                                         \\ \hline
CelebA        & 8                 & 23.97                     & 0.831                     & 0.0231                                         \\ \hline
CelebA        & 16                & 21.211                    & 0.743                     & 0.0233                                         \\ \hline
CelebA        & 32                & 17.848                    & 0.601                     & 0.0232                                         \\ \hline
CIFAR10       & 2                 & 30.627                    & 0.938                     & 0.1019                                         \\ \hline
CIFAR10       & 4                 & 24.975                    & 0.843                     & 0.1055                                         \\ \hline
CIFAR10       & 8                 & 21.008                    & 0.7                       & 0.1055                                         \\ \hline
CIFAR10       & 16                & 18.564                    & 0.576                     & 0.1063                                         \\ \hline
CIFAR10       & 32                & 16.928                    & 0.477                     & 0.1073                                         \\ \hline
CIFAR10(Gray) & 2                 & 31.14                     & 0.912                     & 0.3108                                         \\ \hline
CIFAR10(Gray) & 4                 & 25.336                    & 0.781                     & 0.3312                                         \\ \hline
CIFAR10(Gray) & 8                 & 21.215                    & 0.598                     & 0.3295                                         \\ \hline
CIFAR10(Gray) & 16                & 18.81                     & 0.463                     & 0.3301                                         \\ \hline
CIFAR10(Gray) & 32                & 16.921                    & 0.353                     & 0.3318                                         \\ \hline
Bigset        & 2                 & 38.114                    & 0.973                     & 0.0240                                         \\ \hline
Bigset        & 4                 & 33.706                    & 0.938                     & 0.0243                                         \\ \hline
Bigset        & 8                 & 30.003                    & 0.883                     & 0.0244                                         \\ \hline
Bigset        & 16                & 27.201                    & 0.831                     & 0.0249                                         \\ \hline
Bigset        & 32                & 24.191                    & 0.758                     & 0.0249                                         \\ \hline
Bigset(Gray)  & 2                 & 38.697                    & 0.942                     & 0.0720                                         \\ \hline
Bigset(Gray)  & 4                 & 34.293                    & 0.874                     & 0.0728                                         \\ \hline
Bigset(Gray)  & 8                 & 30.449                    & 0.781                     & 0.0726                                         \\ \hline
Bigset(Gray)  & 16                & 27.393                    & 0.692                     & 0.0741                                         \\ \hline
Bigset(Gray)  & 32                & 24.426                    & 0.592                     & 0.0744                                         \\ \hline
\end{tabular}
\vspace{4mm}
\caption{Benchmark results of NLR-CS}
\label{nlrcsresults}
\end{table*}
\end{document}